\documentclass[]{elsarticle}
\usepackage{amsmath}
\usepackage{epsfig}
\usepackage{epstopdf}
\usepackage{graphicx}

\begin{document}
\begin{frontmatter}
\title{A-Ward$_{p\beta}$: Effective hierarchical clustering using the Minkowski metric and a fast \textit{k}-means initialisation\footnote{This is an accepted manuscript in Information Sciences, Elsevier.\\
\textcopyright 2016.  This manuscript version is made available under the CC-BY-NC-ND 4.0 license http://creativecommons.org/licenses/by-nc-nd/4.0/}}
%

\author[mymainaddress]{Renato Cordeiro de Amorim\corref{mycorrespondingauthor}}
\cortext[mycorrespondingauthor]{Corresponding author at School of Computer Science, University of Hertfordshire, College Lane Campus, Hatfield AL10 9AB, UK. Phone:+44 01707 284345 Fax:+44 01707 284115.}
\ead{r.amorim@herts.ac.uk}
\author[mysecondaryaddress]{Vladimir Makarenkov}
\address[mymainaddress]{School of Computer Science, University of Hertfordshire, College Lane Campus, Hatfield AL10 9AB, UK.}
\address[mysecondaryaddress]{D\'epartement d'Informatique, Universit\'e du Qu\'ebec \`a Montr\'eal, C.P. 8888 succ. Centre-Ville, Montreal (QC) H3C 3P8 Canada.}
\ead{makarenkov.vladimir@uqam.ca}

\author[mythirdaddress,myfourthaddress]{Boris Mirkin}
\address[mythirdaddress]{Department of Data Analysis and Machine Intelligence, National Research University Higher School of Economics, Moscow, Russian Federation.}
\address[myfourthaddress]{Department of Computer Science and Information Systems, Birkbeck University of London, Malet Street, London WC1E 7HX, UK.}
\ead{bmirkin@hse.ru}

\begin{abstract}
In this paper we make two novel contributions to hierarchical clustering. First, we introduce an anomalous pattern initialisation method for hierarchical clustering algorithms, called A-Ward, capable of substantially reducing the time they take to converge. This method generates an initial partition with a sufficiently large number of clusters. This allows the cluster merging process to start from this partition rather than from a trivial partition composed solely of singletons. 

Our second contribution is an extension of the Ward and Ward$_p$ algorithms to the situation where the feature weight exponent can differ from the exponent of the Minkowski distance. This new method, called A-Ward$_{p\beta}$, is able to generate a much wider variety of clustering solutions. We also demonstrate that its parameters can be estimated reasonably well by using a cluster validity index.

We perform numerous experiments using data sets with two types of noise, insertion of noise features and blurring within-cluster values of some features. These experiments allow us to conclude: (i) our anomalous pattern initialisation method does indeed reduce the time a hierarchical clustering algorithm takes to complete, without negatively impacting its cluster recovery ability; (ii) A-Ward$_{p\beta}$ provides better cluster recovery than both Ward and Ward$_p$.
\end{abstract}
\begin{keyword}
Initialisation algorithm \sep Minkowski metric \sep hierarchical clustering \sep feature weighting.
\end{keyword}
\end{frontmatter}
\section{Introduction}
Clustering algorithms are a popular choice when tackling problems requiring exploratory data analysis. In this scenario, analysts can draw conclusions about data at hand without having information regarding the class membership of the given entities. Clustering algorithms aim at partitioning a given data set $Y$ into $K$ homogeneous clusters $S=\{S_1, S_2, ..., S_K\}$ without requiring any label learning process. These algorithms summarise information about each cluster by producing $K$ centroids, often called prototypes, $C=\{c_1, c_2, ..., c_K\}$. The ability to partition data and to provide information about each part has made the application of clustering popular in many fields, including: data mining, computer vision, security, and bioinformatics \cite{jain2010data,mirkin2012clustering,leiva2013warped,steinley2006k,makarenkov2001optimal,maldonado2015kernel}.

There are various approaches to data clustering, with algorithms often divided into partitional and hierarchical. Originally, partitional algorithms produced only disjoint clusters so that each entity $y_i \in Y$ was assigned to a single cluster $S_k$. This hard clustering approach has been variously extended to fuzzy sets \cite{zadeh1965fuzzy}. Fuzzy clustering allows a given entity $y_i \in Y$ to belong to each cluster $S_k \in S$ with different degrees of membership. There are indeed a number of partitional algorithms, with \textit{k}-means \cite{ball1967clustering,macqueen1967some} and fuzzy c-means \cite{bezdek1984fcm} being arguably the most popular under the hard and fuzzy approach, respectively.

Hierarchical algorithms provide additional information about data. They generate a clustering $S$ and related set of centroids $C$, very much like partitional algorithms, but they also give information regarding the relationships among clusters. This information comes as a nested sequence of partitions. This tree-like relationship can be visualized with a dendrogram (i.e., an ultrametric tree). In this type of clustering, an entity $y_i \in Y$ may be assigned to more than one cluster as long as the clusters are related and the assignment occurs at different levels of the hierarchy. 

Hierarchical algorithms can be divided into agglomerative and divisive \cite{mirkin2012clustering}. Agglomerative algorithms follow a bottom-up approach. They start by setting each entity $y_i \in Y$ as the centroid of its own cluster (singleton). Pairs of clusters are then merged stepwise until all the entities have been collected in the same cluster, or until a pre-specified number of clusters is found. Divisive algorithms do the opposite by following a top-down approach. 

There is indeed a wide variety of algorithms to apply when using hierarchical clustering. The Ward method \cite{ward1963hierarchical} is one of the most popular hierarchical algorithms. It follows the agglomerative approach, merging at each iteration the two clusters that minimise the within-cluster variance. This variance is measured as a weighted sum of squares, taking into account the cardinality of each cluster, and leading to the cost function as follows:

\begin{equation}
\label{Eq:Ward}
Ward(S_a, S_b) = \frac{N_a N_b}{N_a+N_b} \sum_{v=1}^V (c_{av} - c_{bv})^2,
\end{equation}
where $V$ is the number of features used to describe each entity $y_i \in Y$. $N_a$ and $c_a$ represent the cardinality and centroid of cluster $S_a \in S$, respectively. Similarly, we have $N_b$ and $c_b$ for cluster $S_b \in S$. The fraction in (\ref{Eq:Ward}) ensures that if two pairs of clusters are equally apart, those of lower cardinalities are merged first.

Previously, we extended the traditional Ward algorithm by introducing Ward$_p$ \cite{de2015feature}. Our algorithm applies cluster dependent feature weights and extends the squared Euclidean distance in (\ref{Eq:Ward}) to the $p$-th power of the weighted Minkowski distance. With these we: (i) ensure that relevant features have a higher impact in the clustering than those that are less relevant; (ii) can set the distance bias to other shapes than that of a spherical cluster, a problem traditionally addressed by methods following model-based clustering \cite{fraley1998many}.

The contribution of this paper is two-fold. First, we introduce what we believe to be the first non-trivial initialisation method for a hierarchical clustering algorithm. Our method generates an initial partition with a sufficiently large  number of clusters. Then, the merging process applies starting from this partition rather than from the singletons. In this way, the running time of a given hierarchical clustering algorithm is substantially reduced. Second, we advance hierarchical clustering by introducing A-Ward$_{p\beta}$, an extension of Ward$_p$ to the situation in which our initialisation method applies and the feature weight exponent can differ from the exponent of the Minkowski distance. We give a rule for choosing these two exponents for any given data set. We run numerous computational experiments, with and without noise in data sets. 

It is worth noting that the ``noise'' in this paper has nothing to do with the conventional meaning of measurement errors, which are usually modelled by an additive or multiplicative Gaussian distribution affecting every data entry. Here, the noise is modelled by either of two ways: (1) inserting additional random noise features, and (2) blurring some features within some clusters. We establish that: (i) the initial clustering generated by our method does decrease the time a hierarchical clustering algorithm takes to complete; (ii) A-Ward$_{p\beta}$ provides a better cluster recovery under different noise models, than either the Ward or the Ward$_p$ algorithms, especially for noisy data.

We direct readers interested to know more of feature weighting in the square-error clustering to reviews such as \cite{kriegel2009clustering}, and references within.


\section{Ward clustering using anomalous patterns}
\subsection{Ward and anomalous pattern Ward}

\textit{K}-means is arguably the most popular partitional clustering algorithm \cite{jain2010data,steinley2006k}. It can be considered an analogue to the general expectation-maximisation algorithm (EM) \cite{dempster1977maximum}. Note, however, that EM recovers a mixed distribution density function, whereas \textit{k}-means just finds a set of non-overlapping clusters and their centres. \textit{K}-means alternatingly minimises the within cluster sum of squares:
\begin{equation}
\label{Eq:KMeans}
W(S,C) = \sum_{k=1}^K \sum_{y_i \in S_k} \sum_{v=1}^V (y_{iv} - c_{kv})^2
\end{equation}
to obtain a partition of the given set of $N$ entities in a set of non-overlapping clusters $S_k \in S$, each represented by its centroid $c_k$, $k=1,2,..., K$.
This minimisation is usually done by following the three straightforward steps: (i) set the coordinates of each centroid $c_k \in C$ to a randomly chosen entity $y_i \in Y$; (ii) assign each entity $y_i \in Y$ to the cluster $S_k$ whose centroid $c_k$ is the nearest to $y_i$; (iii) update each centroid $c_k \in C$ to the component-wise mean of $y_i \in S_k$. Steps (ii) and (iii) are repeated until convergence.

The popular Ward agglomeration algorithm \cite{ward1963hierarchical} uses the same criterion to build a sequence of partitions, each obtained by merging two  clusters $S_a$ and $S_b$, that are the nearest according to (\ref{Eq:Ward}),
%
sometimes referred to as Ward distance between the clusters. 
The Ward algorithm can be formulated as follows:\\

\textbf{Ward agglomeration algorithm}

\begin{enumerate}
\setlength{\itemsep}{-1pt}
\item \textit{Initial Setting}. Set the initial number of clusters $K=N$ and the related singleton clustering $S=\{S_1, S_2, ..., S_{N}\}$ in which every cluster consists of a single element of the data set, so that its centroid is the same element.
\item \textit{Merge clusters}. Using (\ref{Eq:Ward}), find the two nearest clusters $\{S_a, S_b\} \subseteq S$. Merge $S_a$ and $S_b$, creating a new cluster $S_{ab}$. Remove all references to $S_a$, $S_b$, $c_a$, and $c_b$.
\item \textit{Centroid update}. Set the centroid of $S_{ab}$ to the component-wise mean of $y_i \in S_{ab}$.
\item \textit{Stop condition}. Reduce $K$ by 1. If $K>1$ or if $K$ is still higher than the desired number of clusters, go back to Step 2.
\end{enumerate}
Both \textit{k}-means and the Ward method minimise the sum of squared errors, but there are considerable differences in their time-complexity. \textit{K}-means has a linear time-complexity on the number of entities, of $\mathcal{O}(NIKV)$ \cite{tan2006introduction}, where $I$ is the number of iterations it takes to converge and $K$ is the given number of classes. The number of iterations, $I$, is often small and can be reduced even further if \textit{k}-means is supplied with good initial centroids. 

The first implementations of Ward had the time complexity of $\mathcal{O}(N^3)$ and $\mathcal{O}(N^2log^2(N))$ \cite{eppstein2000fast} when a dissimilarity matrix between entities of size $(N\times N)$ was used as input. However, the optimal implementation of Ward, which is due to the development of the nearest neighbour chain and reciprocal nearest neighbour algorithms \cite{Juan1982Programme, murtagh1983survey}, is in $\mathcal{O}(N^2)$. For instance, Murtagh \cite{murtagh1985multidimensional} and, more recently, Murtagh and Legendre \cite{murtagh2014ward} discussed in detail the nearest neighbour chain algorithm using either ``stored data'' or ``stored dissimilarities'' implementations, leading to $\mathcal{O}(N^2)$ computational complexity of Ward. Nowadays, optimal implementations of the Ward algorithm became standard and are widely used in the popular software packages, such as R \cite{r_stats}, Clustan \cite{Clustan} or MATLAB \cite{matlab_stats}.

There are many initialisation methods for \textit{k}-means \cite{bradley1998refining,pena1999empirical,steinley2006k}. Milligan \cite{milligan1980} pointed out that the results of \textit{k}-means heavily depend on initial partitioning. He suggested that a good final clustering can be obtained using Ward's hierarchical algorithm to initialise it, which was confirmed later in computational experiments (see, for example, \cite{steinley2006k}). There are also other examples of using hierarchical clustering to initialise \textit{k}-means \cite{su2007search,cao2009initialization,celebi2012deterministic}. Conversely, \textit{k}-means is beneficial as a device for carrying out divisive clustering, see, for example, what is referred to as the ``bisecting \textit{k}-means'' \cite{steinbach2000comparison, mirkin2012clustering}. The author of the Clustan package \cite{Clustan}, David Wishart, was first to propose the \textit{k}-means Cluster Model Tree method which allows one to summarize a \textit{k}-means cluster solution by a hierarchy. For instance, a mini-tree for each \textit{k}-means cluster, showing how the entities combine within this cluster, can be constructed and visualized using Clustan \cite{Clustan}. However, to the best of our knowledge, the problem of accelerating agglomerative clustering using \textit{k}-means has not been addressed so far.

This problem is related to the problem of pre-selecting the right value for the number of clusters $K$ when running \textit{k}-means. Such a pre-selected number of clusters should be greater than the number of expected clusters, but not too much. We propose using the method of intelligent \textit{k}-means (\textit{ik}-means) \cite{chiang2010intelligent, mirkin2012clustering} for this purpose. This method finds and removes ``anomalous'' clusters, one-by-one, from the data set, so that the number of these clusters is not pre-specified but rather obtained according to the data set structure by using a threshold $\theta$ that is the minimum number of entities required to form a cluster. When this threshold is set to 1, the number of anomalous clusters has been experimentally found to be always larger than the number of generated clusters. The \textit{ik}-means algorithm finds the current anomalous cluster $S_t$ and respective centroid $c_t$ by alternatingly minimising:
\begin{equation}
	W(S_t,c_t) = \sum_{i \in S_t} d(y_i,c_t) + \sum_{i \notin S_t} d(y_i,0),
\end{equation}
where $d(y_i,c_t)$ is the squared Euclidean distance between entity $y_i$ and centroid $c_t$, and $d(y_i,0)$ is the squared Euclidean distance between entity $y_i$ and the centre of the data set $Y$. The algorithm then removes $S_t$ from the data set and re-applies the process to the remaining entities as explained below. Thus, the number of anomalous clusters, $K^*$, is our criterion for a fast preliminary estimation of the true number of clusters in the data set.\\

\textbf{Anomalous cluster identification algorithm (\textit{ik}-means)}

\begin{enumerate}
\setlength{\itemsep}{-1pt}
\item \textit{Initial setting.} Set the user-defined $\theta$. Set the centroid $c_Y$ to be the component-wise mean of $y_i \in Y$.
\item \textit{Tentative centroid.} Set $S_t=\emptyset$. Set $c_t$, a tentative centroid, to coincide with the entity $y_i \in Y$ that is farthest from $c_Y$ according to the squared Euclidean distance.
\item \textit{Entity assignment.} Assign each entity $y_i \in Y$ to either $c_t$ or to $c_Y$ depending on which is the nearest. Those assigned to $c_t$ form the cluster $S_t$. If there are no changes in $S_t$, go to Step 5.
\item \textit{Centroid update.} Update $c_t$ to the component-wise mean of $y_i \in S_t$. Go to Step 3.
\item \textit{Save centroid.} If $|S_t|\geq \theta$, include $c_t$ into $C$.
\item \textit{Remove clusters.} Remove each $y_i \in S_t$ from $Y$. If $|Y|>0$, go to Step 2.
\item \textit{Cluster.} Run \textit{k}-means on the original data set $Y$, using as initial centroids those in $C$.
\end{enumerate}
The above is a rather successful initialisation for \textit{k}-means \cite{chiang2010intelligent}. We use it as a base for our anomalous pattern initialisation method for agglomerative clustering algorithms described later in this section.

The traditional Ward algorithm starts from a trivial clustering $S=\{S_1, S_2, ..., S_N\}$ in which every cluster is a singleton. The sole purpose of $S$ is to serve as a base for the clustering generated in the next iteration of Ward. Obviously, this trivial set is useless to any data analyst. With the above in mind, one could wonder whether the clustering generated in the next iteration of Ward, that with $N-1$ clusters, would be of any interest to a data analyst. This will be a clustering in which only one of the $N-1$ clusters is not a singleton. Of course, we cannot state if it is of any interest or not because the degree of usefulness of such clustering is problem-dependent. However, classifying $N$ entities into $N-1$ classes would be trivial in most of the practical situations.

If neither $N$ nor $N-1$ clusters would constitute a useful clustering, we could challenge the usefulness of the solution with $N-2$ clusters and so on. Clearly, at some stage we will reach a number of clusters, $K^*$, that leads to a useful clustering in terms of partitions. $K^*$ is not a reference to the true number of clusters in $Y$, even if such number is known. Instead, $K^*$ represents the number of clusters in which the data begins to manifest some form of cluster structure. Since in this paper we follow the agglomerative approach, $K^*$ can be also viewed as the maximum number of anomalous patterns in $Y$.

Above, we described the \textit{ik}-means. This is an algorithm able to find anomalous patterns in a data set, as well as the related partitions. The maximum number of anomalous patterns $K^*$ in $Y$ is given by \textit{ik}-means when the parameter $\theta$ is set to 1. This setting leads to two important points: (i) it allows for the possibility of singletons; (ii) $K^*$ is greater than the true number of clusters in $Y$.

Ideally, Ward should be initialised with $K^*$ and the related clustering generated by \textit{ik}-means. The point (i) is important because $Y$ may be a sample of a larger real-world data set. It is possible that the larger data set contains a cluster $|S_k|>1$ for which the sample $Y$ contains a single entity $\in S_k$. Moreover, since $K^*$ is an overestimation of the true number of clusters in $Y$ (ii), our version of Ward will generate a tree structure from $K^*$ until the true number of clusters is found. If the latter is unknown, we can generate a binary hierarchy beginning with $K=K^*$ and finishing with $K=2$.

The main objective of our method is to reduce the number of steps Ward takes to complete, and by consequence, the time required for its execution. The results we present later in this section show that the gain in running time provided by this strategy can be very significant (see Figures \ref{Fig:Ward_Time} and \ref{Fig:TTWard_Time}). Now we can formalise Ward with anomalous pattern initialisation, further on referred to as A-Ward, as follows:\\

\textbf{A-Ward algorithm}
 
\begin{enumerate}
\setlength{\itemsep}{-1pt}
\item \textit{Initial Setting}. Set $\theta = 1$. Obtain the initial number of clusters $K=K^*=|C|$ and the related clustering $S=\{S_1, S_2, ..., S_{K}\}$ by running \textit{ik}-means on $Y$.
\item \textit{Merge clusters}. Using (\ref{Eq:Ward}), find the two closest clusters $\{S_a, S_b\} \subseteq S$. Merge $S_a$ and $S_b$, creating a new cluster $S_{ab}$. Remove all references to $S_a$, $S_b$, $c_a$, and $c_b$.
\item \textit{Centroid update}. Set the centroid of $S_{ab}$ to the component-wise mean of $y_i \in S_{ab}$.
\item \textit{Stop condition}. Reduce $K$ by 1. If $K>2$ or if $K$ is still higher than the desired number of clusters, go back to Step 2.
\end{enumerate}

\subsection{Comparing Ward and A-Ward}
\label{Sec:ExperimentsSetting}

When defining the A-Ward algorithm, we intended to define a method that has a similar cluster recovery capability with Ward, while being somewhat faster. To test a new clustering method, it is quite natural to define a collection of data sets with a predefined cluster structure, which is the easiest to achieve by generating synthetic data sets.  Using synthetic data with and without noise, we can apply both Ward and A-Ward clustering algorithm and assess both the speed and the level of cluster recovery. To measure the level of cluster recovery, we compare the cluster-found partition with the generated reference partition by using the conventional Adjusted Rand Index \cite{hubertarabie1985rand}. This popular index is the corrected for chance version of the Rand index: 

\begin{equation}
\label{Eq:ARI}
ARI = \frac{ \sum_{ij} \binom{n_{ij}}{2} - [\sum_i \binom{a_i}{2} \sum_j \binom{b_j}{2}] / \binom{n}{2} }{ \frac{1}{2} [\sum_i \binom{a_i}{2} + \sum_j \binom{b_j}{2}] - [\sum_i \binom{a_i}{2} \sum_j \binom{b_j}{2}] / \binom{n}{2} },
\end{equation}
where $n_{ij}=|S_i \cap S_j|$, $a_i = \sum_{j=1}^K |S_i \cap S_j|$ and $b_i = \sum_{i=1}^K |S_i \cap S_j|$. The range of (\ref{Eq:ARI}) is within the interval from -1 to 1. ARI reaches 1 if and only if the two compared partitions coincide, i.e., $S_p=S_q$.

We begin by generating 20 synthetic data sets for each of the configurations 1000x6-3, 1000x12-6 and 1000x20-10 (for details see Table \ref{Tab:GMMs}). In these data sets, all clusters are spherical. That is,  each cluster is generated from a Gaussian distribution whose covariance matrix is diagonal with the same diagonal value $\sigma^2$ generated randomly between $0.5$ and $1.5$. Each of the centroid components was generated independently using the standard normal distribution $N(0,1)$. The cardinality of each cluster was selected from a uniform distribution, with the constraint that it should have at least 20 entities.

Then we introduced noise in these data sets according to either of the two following noise generation models:

\begin{enumerate}
\item {\bf Noise model 1: Random feature to be inserted.}
A noise feature is generated according to a uniform distribution in the range between the minimum and maximum values in the data set.
\item {\bf Noise model 2: Blurring a cluster over a feature.}
Any feature in a generated data set contains $K$ cluster-specific fragments. By randomly selecting a feature and cluster, such a fragment is substituted by a uniform random noise. 
\end{enumerate}
The noise model 1 addresses the issue of generic clustering methods based on the least-squares criterion (\ref{Eq:KMeans}): they cannot distinguish between useful and inadequate features. It has been used in \cite{cordeiro2011minkowski, de2015feature,de2015recovering,de2016applying} to test the weighted feature versions of $k$-means and Ward algorithms; those showed good cluster recovery properties against such noise features. The noise model 2 is novel. It is supposed to be applied for testing the ability of clustering algorithms to perform under the cluster-specific noise. In practice this type of noise can be found in various fields, including computer vision \cite{freytag2012efficient}, financial economics \cite{wilcox2014hierarchical} and genomics\cite{monni2009stochastic}.

We added 50\% of noise data to each of the original data sets according to each of the above-defined noise models. For example, each of the 20 data sets generated according to the configuration 1000x12-6 contains 12 original features; six noise features have been inserted into each of them (leading to a total of 18 features). We refer to this new configuration as 1000x12-6+6NF, where NF stands for "noise feature". Similarly, 50\% of all the $KV$ cluster-specific fragments have been blurred according to the noise model 2, which is denoted here as 1000x12-6 50\%N.

Our simulations were carried out using a 64-bit computer equipped with an Intel i5-4690T CPU, running at 2.5GHz, and 8Gb of RAM. Our algorithms were implemented using MATLAB R2013 running on Linux (Ubuntu). We did not use the partially pre-compiled MATLAB's \texttt{linkage} function as it would introduce bias to our experiments.

\begin{table}\small
\caption{The nine cluster structure configurations used in  simulations.}
\begin{center}
\tabcolsep=0.11cm
\begin{tabular}{lccccc}
&Entities&Features&Clusters&Noise&Cluster-specific\\
&&&&features&noise (\%)\\
\cline{2-6}
1000x6-3&1000&6&3&0&0\\
1000x6-3 +3NF&1000&6&3&3&0\\
1000x6-3 50\%N&1000&6&3&0&50\\
\hline
1000x12-6&1000&12&6&0&0\\
1000x12-6 +6NF&1000&12&6&6&0\\
1000x12-6 50\%N&1000&12&6&0&50\\
\hline
1000x20-10&1000&20&10&0&0\\
1000x20-10 +10NF&1000&20&10&10&0\\
1000x20-10 50\%N&1000&20&10&0&50\\
\end{tabular}
\end{center}
\label{Tab:GMMs}
\end{table}

\begin{figure}[t]
  \caption{Time in seconds the conventional Ward algorithm takes to complete.}
  \centering
    \includegraphics[width=1\textwidth]{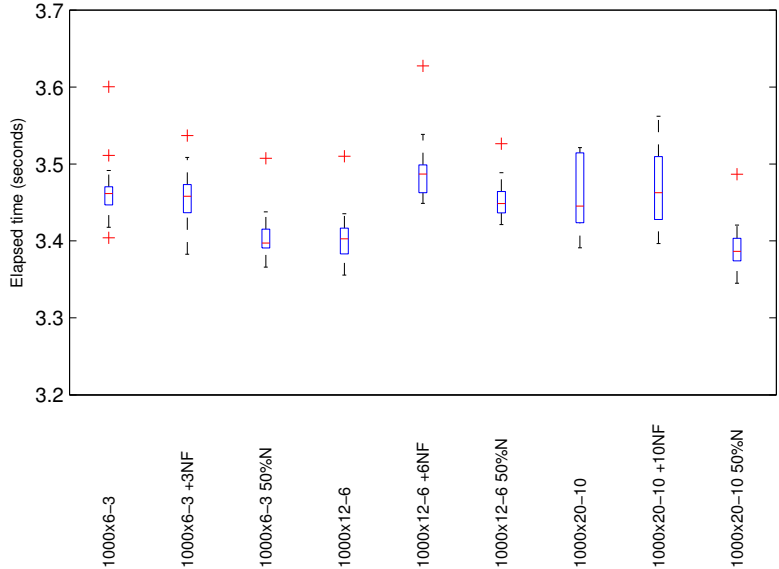}
\label{Fig:Ward_Time}
\end{figure}

\begin{figure}[b]
  \caption{Time in seconds the A-Ward algorithm takes to complete.}
  \centering
    \includegraphics[width=1\textwidth]{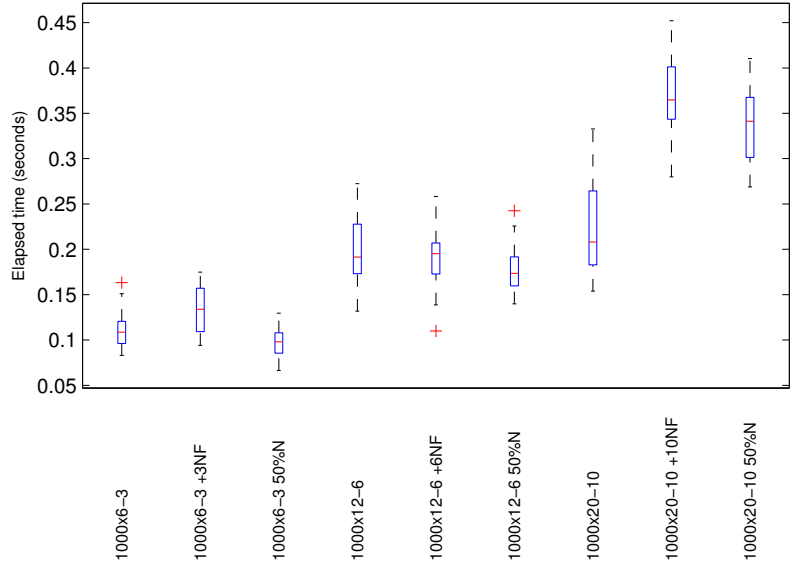}
\label{Fig:TTWard_Time}
\end{figure}

The results of running Ward and A-Ward over the $180=9\times 20$ generated data sets confirm our assumptions:
\begin{enumerate}
\item A-Ward is significantly  faster than Ward (see Figures \ref{Fig:Ward_Time} and \ref{Fig:TTWard_Time} demonstrating time box-plots for each of the data configurations);
\item A-Ward and Ward have similar cluster recovery capabilities over each of the data set configurations (see Table \ref{Tab:GMM_Baseline_Ward_TTWard}).
\end{enumerate}

Table \ref{Tab:GMM_Baseline_Ward_TTWard} reports the number of anomalous clusters $K^*$ found by \textit{ik}-means. The presented results suggest that this number is indeed greater than the number of generated clusters. We also computed the average ARI values between the solutions provided by Ward and A-Ward (see the last two columns in Table \ref{Tab:GMM_Baseline_Ward_TTWard}). This additional ARI is close to the ARI between the solutions provided by both algorithms and the known truth for data sets without noise. The ARI values increase with the increase in the number of features, still for data not affected by noise. 

For data sets including noise, the trend is quite the opposite. In these cases, we can conclude that the solutions yielded by Ward and A-Ward diverge, and this divergence can be very significant as the quantity of noise increases. Clearly, both Ward and A-Ward appear to be absolutely impractical in the presence of noise.

The optimal time complexity of the Ward algorithm is $\mathcal{O}(N^2V)$ given that an object-to-feature $(N\times V)$ data matrix is used as input \cite{murtagh2014ward}. Our anomalous pattern method initialises Ward with $K^*$ clusters instead of $N$, leading to the time complexity of the remaining Ward operations of $\mathcal{O}(K^{*2}V)$, i.e., after initialisation with \textit{ik}-means. The average values of $K^*$ over the processed data sets (see Table \ref{Tab:GMM_Baseline_Ward_TTWard}) vary from 19.90 to 49.95. Obviously, the initialisation stage of A-Ward has also a computational cost expressed via the time complexity of \textit{ik}-means, which is of $\mathcal{O}(NK^*IV)$, where $I$ is the number of iterations \textit{ik}-means takes to converge. Thus, we can claim, after dividing the involved time complexities by $V$, that our A-Ward algorithm decreases the amount of time that traditional Ward takes to complete as long as: $\mathcal{O}_k(NIK^*)<\mathcal{O}_w(N^2-K^{*2})$, where $\mathcal{O}_k$ is the upper bound of \textit{ik}-means and $\mathcal{O}_w$ is the upper bound of Ward.

\begin{figure}[ht]
  \caption{ARI of Ward (left of each pair of boxes) and A-Ward (right of each pair of boxes) for data sets in which 20\% of entities were substituted by within-domain uniformly random noise. The ARI was calculated without taking the substituted entities into account.}
  \centering
    \includegraphics[width=1.2\textwidth]{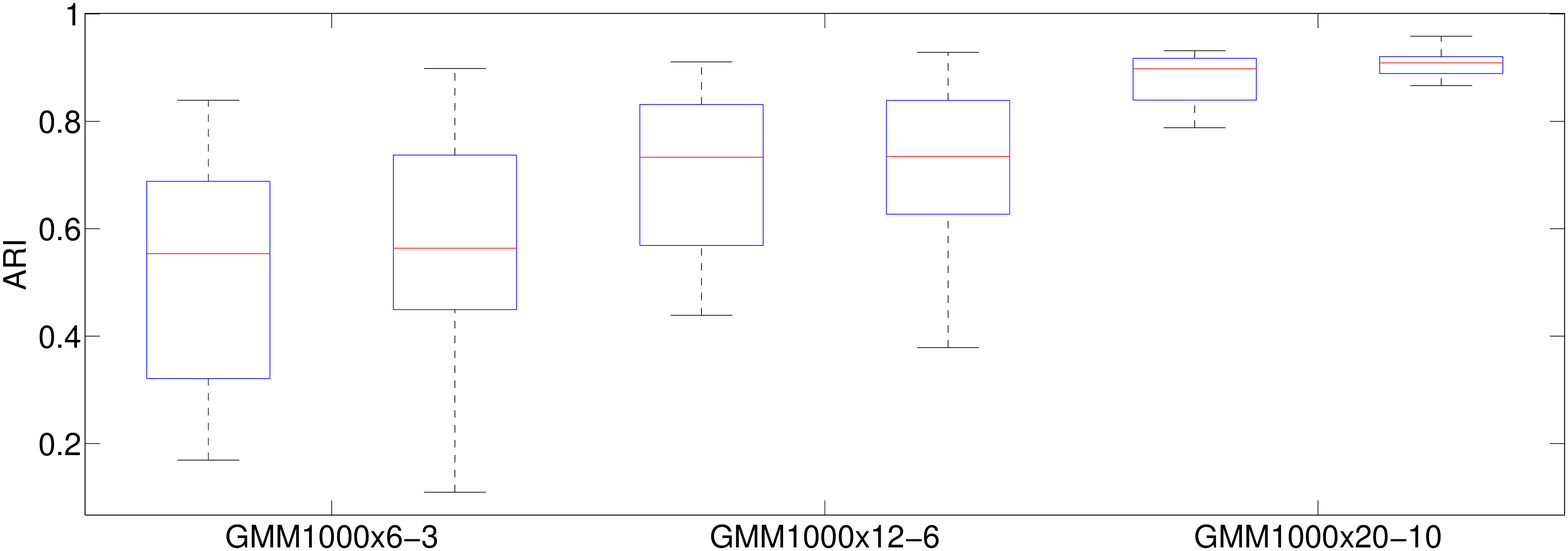}
\label{Fig:ARIWard_AWard}
\end{figure}

\begin{table}
\caption{The average ARI, its standard deviation and the number of pre-selected clusters $K^*$ for the  Ward and A-Ward algorithms obtained over 20 synthetic data sets for each of the nine parameter configurations.}
\begin{center}
\tabcolsep=0.11cm
\begin{tabular}{lccccccccccc}
&\multicolumn{2}{c}{Ward}&&\multicolumn{5}{c}{A-Ward}&&\multicolumn{2}{c}{Ward/A-Ward}\\
\cline{5-9}
&&&&\multicolumn{2}{c}{ARI}&&\multicolumn{2}{c}{$K^*$}&&\multicolumn{2}{c}{ARI}\\
\cline{2-3}
\cline{5-6}
\cline{8-9}
\cline{11-12}
&avg&sd&&avg&sd&&avg&sd&&avg&sd\\
1000x6-3&0.5448&0.231&&0.5285&0.197&&19.90&2.245&&0.5217&0.204\\
1000x6-3 +3NF&0.0400&0.109&&0.0501&0.132&&22.70&2.934&&0.3046&0.153\\
1000x6-3 50\%N&0.0545&0.090&&0.0877&0.108&&20.20&2.262&&0.2910&0.157\\
\hline
1000x12-6&0.6929&0.166&&0.7102&0.188&&33.55&6.082&&0.6669&0.185\\
1000x12-6 +6NF&0.1375&0.130&&0.1267&0.123&&26.20&4.937&&0.2093&0.079\\
1000x12-6 50\%N&0.1276&0.089&&0.1208&0.078&&28.65&4.221&&0.2096&0.057\\
\hline
1000x20-10&0.8998&0.060&&0.9058&0.061&&36.40&7.229&&0.8704&0.078\\
1000x20-10 +10NF&0.2418&0.084&&0.2326&0.096&&49.75&8.226&&0.1871&0.055\\
1000x20-10 50\%N&0.1360&0.048&&0.1283&0.043&&49.95&8.636&&0.1617&0.035\\
\end{tabular}
\end{center}
\label{Tab:GMM_Baseline_Ward_TTWard}
\end{table}
Usually, hierarchical algorithms are sensitive to perturbations that affect all entities in data sets. Thus, we carried out experiments to determine the impact of our initialisation method in such a case. To do so we substituted 20\% of the entities, rather than features, of each data set by uniformly random noise. We then calculated the ARI between the clusterings obtained with Ward and A-Ward to the known truth, without taking the substituted entities into account. We performed this set of experiments on data sets without any additional noise. The obtained results are presented in Figure \ref{Fig:ARIWard_AWard}. We can observe that A-Ward produces the largest ARI range for the 1000x6-3 and 1000x12-6 data set configurations. However, A-Ward provides the highest first and third quartiles, as well as the median, in all the cases.

\subsection{Case study}
In this subsection we present an example of application of our A-Ward algorithm. Our main objective is to demonstrate that the \textit{ik}-means initialisation used by A-Ward does not negatively impact its ability to recover clusters. To do so, we considered the popular Zoo data set, which can be found in the UCI machine learning repository \cite{Lichman:2013}.

Species hierarchies are usually relatively easy to understand and interpret. The Zoo data set contains 101 entities, described over 16 features, and partitioned into seven clusters. We have treated all features as numeric and standardised them as follows:
\begin{equation}
\label{Eq:Stand}
y^{\prime}_{iv} = \frac{y_{iv}-\overline{y_v}}{max(y_v)-min(y_v)},
\end{equation}
where $\overline{y_v}$ is the average value of $v$ over all entities in $Y$, and $y^{\prime}_{iv}$ is the standardised value of $y_{iv}$. 

Our choice of standardisation method has two important implications. First, unlike \textit{z}-score it does not favour a unimodal distribution. This is probably easier to explain with an example. Consider a unimodal feature $v_1$ and bimodal feature $v_2$. The standard deviation of $v_2$ is likely to be higher than that of $v_1$, leading to $y^{\prime}_{iv_{2}} < y^{\prime}_{iv_{1}}$. This is particularly problematic because clustering would usually target the groups associated with the modes in $v_2$.

Second, if $v$ is a binary feature its range will be one. This means that the standardised value of $y_{iv}$ is simply $y_{iv}-\overline{y_v}$. With this, features with a higher frequency lead to lower standardised values than features with lower frequencies. For example, the binary features with multiple zero values will have a very significant impact on the clustering process.

\begin{figure}[t]
  \caption{Zoo hierarchy found by our A-Ward algorithm for 28 species of the Zoo dataset (4 species from each of the 7 original Zoo classes were selected
randomly). The species content by class is as follows: Class 1: porpoise, platypus, reindeer, fruitbat; Class 2: dove, gull, swan, rhea; Class 3: pitviper, slowworm, tortoise, tuatara; Class 4: herring, sole, carp, stingray; Class 5: frog1, frog2, newt, toad; Class 6: wasp, honeybee, housefly, gnat;
Class 7: crayfish, seawasp, crab, clam. Red circles in the tree represent 11 clusters found by \textit{ik}-means during the initialization step of A-Ward. Red edges of the hierarchy represent the tree found by A-Ward during its tree building step. Green edges of the hierarchy represent mini-trees found by the conventional Ward algorithm (this step is optional) for the 11 clusters provided by \textit{ik}-means.}
  \centering
    \includegraphics[width=0.9\textwidth]{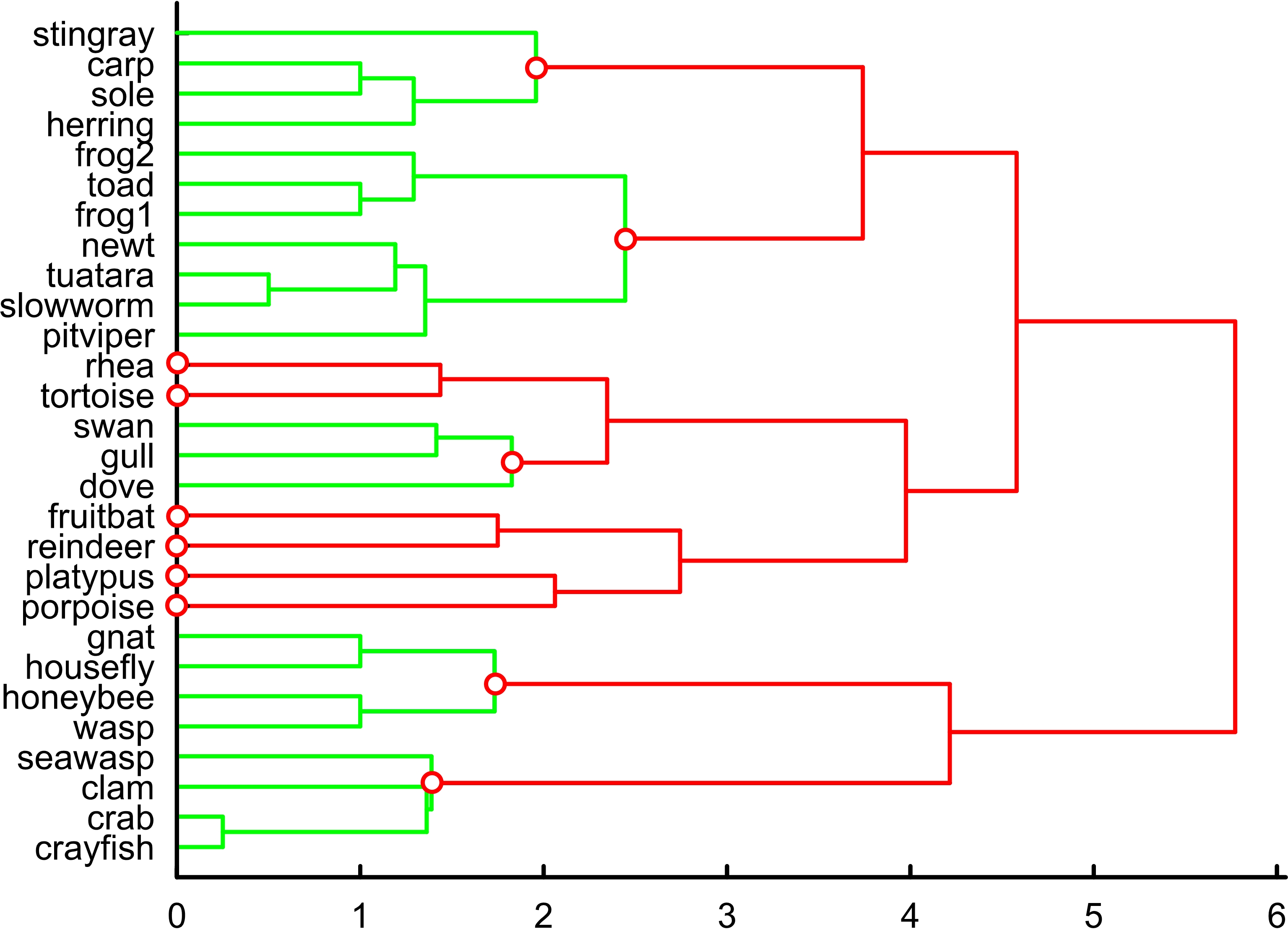}
\label{Fig:Ward_AWard}
\end{figure}

Since the complete Zoo data set is too large to be shown as a single tree, we selected randomly four entities of each of its seven clusters; 28 entities in total. The only misclassified species in the A-Ward hierarchy presented in Figure \ref{Fig:Ward_AWard} is $tortoise$ (from Class 3), which is clustered with the species of Class 2. It is worth noting that a misclassification of $tortoise$ is also characteristic for the traditional Ward algorithm. However, A-Ward produces the top part of the hierarchy, without the computational cost of Ward.

\section{Using the weighted Minkowski distance} 

\subsection{Weighted Minkowski k-means and Ward algorithms}

We previously dealt with noise data sets by introducing the intelligent Minkowski weighted \textit{k}-means algorithm (\textit{imwk}-means)\cite{cordeiro2011minkowski}. This algorithm minimises the following objective function:
\begin{equation}
\label{Eq:MWKMeans}
W(S,C,w) = \sum_{k=1}^K \sum_{y_i \in S_k} \sum_{v=1}^V w_{kv}^{p}|y_{iv} - c_{kv}|^p,
\end{equation}
where $p$ is a user-defined exponent related to  what can be called the curvature bias. Assuming a two-dimensional space (for an easier visualisation), the bias at $p=1$, $p=2$, and $p\rightarrow\infty$ is towards diamonds, circles and squares, respectively. 

The \textit{imwk}-means criterion clearly sets the exponent of the distance and the feature weight to the same value, $p$. Thus, the feature weights can be seen as re-scaling factors for any value of $p$. These rescaling factors can be used in the data pre-processing stage of a wide variety of tasks in machine learning. For instance, rescaling a data set with these factors increases the likelihood of recovering the correct number of clusters contained in the data \cite{de2015recovering}.

The weight of feature $v$ at cluster $S_k$ is inversely proportional to the dispersion of $v$ at $S_k$ since the first-order necessary minimum conditions  of (\ref{Eq:MWKMeans}) imply that:
\begin{equation}
\label{Eq:MWK_Weight}
w_{kv} = \frac{1}{\sum_{u=1}^V[D_{kv}/D_{ku}]^{1/(p-1)}},
\end{equation}
where $D_{kv}=\sum_{i \in S_k} |y_{iv} -c_{kv}|^p$ is the dispersion of $v$ at $S_k$. The above is true for crisp clustering where each entity $y_i \in Y$ is assigned to one and only one cluster $S_k$, leading to $\sum_{v=1}^V w_{kv}=1$, for $k=1, 2, ..., K$. At $p=1$ the minimum of (\ref{Eq:MWKMeans}) is reached at the median. Moreover, because this criterion has a linear shape at $p=1$, the first-order minimum conditions are not applicable here and, therefore, formula (\ref{Eq:MWK_Weight}) is not applicable either. Thus, we run experiments at $p>1$, only.

Given the success of the above-discussed \textit{imwk}-means algorithm, the agglomerative Ward$_p$ was introduced in \cite{de2015feature}, using a hierarchical clustering heuristic in which cluster-dependent feature weights are determined according to (\ref{Eq:MWK_Weight}). Ward$_p$ is an agglomerative hierarchical clustering algorithm. At each iteration, it merges the two clusters that minimise the following dissimilarity function:

\begin{equation}
\label{Eq:MW_Ward}
Ward_p(S_a, S_b) = \frac{N_a N_b}{N_a + N_b}\sum_{v=1}^V (\frac{w_{av}+w_{bv}}{2})^p |c_{av} - c_{bv}|^p.
\end{equation}
Unlike the distance calculations in \textit{imwk}-means, those of Ward$_p$ are  only between centroids $\{c_a, c_b\} \subseteq C$. Thus, the weight  in (\ref{Eq:MW_Ward}) is the average of $w_{av}$ and $w_{bv}$, each calculated using (\ref{Eq:MWK_Weight}).  Ward$_p$ minimises (\ref{Eq:MW_Ward}) following the steps below:\\

\textbf{Ward$_p$ agglomerative clustering algorithm}

\begin{enumerate}
\setlength{\itemsep}{-1pt}
\item \textit{Initial setting}. Select the value of $p$, starting from a partition consisting of $N$ singleton clusters. Each centroid $c_k \in C$ is set to the corresponding entity $y_i \in Y$. Set $w_{kv} = 1/V$ for $k=1, 2, ..., K$ and $v = 1 , 2, ..., V$.
\item \textit{Merge clusters}. Find the two nearest clusters $\{S_a, S_b\} \subseteq S$ with respect to (\ref{Eq:MW_Ward}). Merge $S_a$ and $S_b$, thus creating a new cluster $S_{ab}$. Remove all references to $S_a$, $S_b$, $c_a$, and $c_b$.
\item \textit{Centroid update}. Set the centroid of $S_{ab}$ to the component-wise Minkowski centre of $y_i \in S_{ab}$.
\item \textit{Weight update}. Using (\ref{Eq:MWK_Weight}) compute weights $w_{kv}$ for $k=1, 2, ..., K$ and $v=1, 2, ..., V$.
\item \textit{Stop condition}. Reduce $K$ by 1. If $K>1$ or if $K$ is still greater than the desired number of clusters, go back to Step 2.
\end{enumerate}

The algorithm Ward$_p$ requires the computation of the Minkowski centre at different values of $p$. This centre can be calculated using a steepest descent method \cite{cordeiro2011minkowski,de2015recovering}.

\subsection{Ward$_{p\beta}$ algorithm initialised with anomalous patterns}
Both \textit{imwk}-means and Ward$_p$ apply the same exponent $p$ to the feature weights and the distance in their respective criteria. There are two major reasons to apply the same exponent. First, by doing so there is a single problem-specific parameter to be defined by the user. Since the optimal value of this parameter is usually unknown to the user, it can be estimated by analysing the clusterings produced at different values of $p$. For instance, the user can carry out Ward$_p$ at $p=1.1, 1.2, ..., 5.0$ and choose as optimal the value of $p$ that optimizes a given cluster validity index. In our previous experiments, we successfully applied the Silhouette width \cite{de2015feature}. Obviously, there are many other cluster validity indices that could be used instead (see a recent survey \cite{arbelaitz2012extensive}).

The second reason is that if the same exponent is employed with the feature weights and the distance, then the weights can be seen as feature rescaling factors. These factors can be used in the data pre-processing stage as an instrument to standardise a data set. For instance, rescaling data sets with these factors improves the likelihood of clustering validity indexes to return the true number of clusters in data sets, particularly in those comprising noise features \cite{de2015recovering}.

The above is helpful when the number of clusters in a data set is unknown. Still, in this paper we deal solely with cluster recovery where the number of clusters is known. Clearly, estimating a single parameter is easier than estimating two. However, by using two exponents we detach the cluster shape from the weight exponent, increasing considerably the variety of clustering possibilities.

Taking all of the above into account, we extend here Ward$_p$ to allow the use of different exponents for the distance and the feature weights. During the initialisation step, our new algorithm, A-Ward$_{p\beta}$, measures the distance between an entity $y_i \in Y$ and the centroid $c_k \in C$ of cluster $S_k$ by:

\begin{equation}
\label{Eq:TwoExponentMinkDist}
d_{p\beta}(y_i, c_k) = \sum_{v=1}^V w_{kv}^{\beta} |y_{iv}-c_{kv}|^p,
\end{equation}
where $p$ and $\beta$ are user-defined parameters. In Section \ref{Sec:Estimating_pb} we introduce a method to estimate good values for these parameters.
Our new algorithm makes use of our anomalous pattern initialisation, this time also applying the weighted Minkowski distance, as presented below:\\

\textbf{Anomalous pattern initialisation for A-Ward$_{p\beta}$ and \textit{imwk}-means$_{p\beta}$}

\begin{enumerate}
\setlength{\itemsep}{-1pt}
\item \textit{Initial setting}. Select the values of $p$ and $\beta$. Set the data centre $c_Y$ to be the component-wise Minkowski centre of $y_i \in Y$.
\item \textit{Tentative centroid}. Set $S_t = \emptyset$. Set $w_{kv}=1/V$ for $k=1, 2$ and $v=1, 2, ..., V$. Set $c_t$, a tentative centroid, to the values of the furthest entity $y_i \in Y$ from $c_Y$ as per (\ref{Eq:TwoExponentMinkDist}).
\item \textit{Entity assignment}. Assign each entity $y_i \in Y$ that is closer to $c_t$ than to $c_Y$ as per (\ref{Eq:TwoExponentMinkDist}) to the cluster $S_t$. If this step produces no changes in $S_t$, go to Step 6.
\item \textit{Centroid update}. Update $c_t$ to the component-wise Minkowski centre of $y_i \in S_t$.
\item \textit{Weight update}. Update the feature weights as per (\ref{Eq:MWK_Weight}). Go to Step 3.
\item \textit{Save parameters}. Include $c_t$ into $C$, and $w$ into $W$.
\item \textit{Remove cluster}. Remove each $y_i \in S_t$ from $Y$. If there are still entities in $Y$, go to Step 2.
\end{enumerate}

We can further minimise the distance between entities and centroids by using centroids $C$ and weights $W$ generated above as starting points for the version of our \textit{imwk}-means$_{p\beta}$ algorithm below: \\

\textbf{\textit{imwk}-means$_{p\beta}$ algorithm}

\begin{enumerate}
\setlength{\itemsep}{0pt}
\item \textit{Initial setting}. Set $K=|C|=K^*$, and $S=\emptyset$.
\item \textit{Entity assignment}. Assign each entity $y_i \in Y$ to the cluster $S_k \in S$ that is represented by the centroid $c_k \in C$ that is the closest to $y_i$ as per (\ref{Eq:TwoExponentMinkDist}). If there are no changes in $S$, go to Step 5.
\item \textit{Centroid update}. Update each centroid $c_k \in C$ to the component-wise Minkowski centre of $y_i \in S_k$.
\item \textit{Weight update}. Update each weight $w_{kv}$ for $k=1, 2, ..., K$ and $v=1, 2, ..., V$ as per (\ref{Eq:MWK_Weight}). Go to Step 2.
\item \textit{Output}. Output the clustering $S$, centroids $C$ and weights $W$.
\end{enumerate}
Upon completion of the algorithm above we obtain a clustering $S$, centroids $C$ and weights $w_{kv}$ for $k=1, 2, ..., K$ and $v=1, 2, ..., V$. As we will show in the following sections, these parameters represent good initial settings for our A-Ward$_{p\beta}$. We use this criterion for building a cluster hierarchy with the following cluster-to-cluster dissimilarity measure:
\begin{equation}
\label{Eq:MW_Wardpb}
Ward_{p\beta}(S_a, S_b) = \frac{N_a N_b}{N_a + N_b}\sum_{v=1}^V (\frac{w_{av}+w_{bv}}{2})^{\beta} |c_{av} - c_{bv}|^p.
\end{equation}
Now we can run the agglomerative A-Ward$_{p\beta}$ algorithm as follows:\\

\textbf{A-Ward$_{p\beta}$ agglomerative algorithm}

\begin{enumerate}
\setlength{\itemsep}{-1pt}
\item \textit{Initial setting}. Take the values of $p$ and $\beta$ used in the
\textit{imwk}-means$_{p\beta}$ algorithm and start from the clustering $S$, centres $C$ and weights $w_{kv}$ provided by \textit{imwk}-means$_{p\beta}$. 
\item \textit{Merge clusters}. Find the two nearest clusters $\{S_a, S_b\} \subseteq S$ with respect to (\ref{Eq:MW_Wardpb}). Merge $S_a$ and $S_b$, thus creating a new cluster $S_{ab}$. Remove all references to $S_a$, $S_b$, $c_a$, and $c_b$.
\item \textit{Centroid update}. Set the centroid of $S_{ab}$ to the component-wise Minkowski centre of $y_i \in S_{ab}$.
\item \textit{Weight update}. Using (\ref{Eq:MWK_Weight}) compute weights $w_{kv}$ for $k=1, 2, ..., K$ and $v=1, 2, ..., V$.
\item \textit{Stop condition}. Reduce $K$ by 1. If $K>1$ or if $K$ is still greater than the desired number of clusters, go back to Step 2.
\end{enumerate}

\subsection{Validation of the A-Ward$_{p\beta}$ algorithm}
 
Analogously to our previous simulation studies \cite{de2016applying,de2015feature}, we first found a set of partitions, each corresponding to a different combination of values of $p$ and $\beta$. The set of all possible values of $p$ and $\beta$ was modelled using a grid of $p$ and $\beta$ values varying from $1.1$ to $5.0$ with the step of $0.1$, as in  \cite{cordeiro2011minkowski}.  We obtained the results for Ward$_p$ by running it with $p=1.1, 1.2, ..., 5.0$ and selecting the clustering with the highest ARI in relation to the known truth. Similarly, the results under Ward$_{p\beta}$ are given with respect to the clusterings with the highest ARI in relation to the known truth. These settings give us an indication of the best possible results we could obtain if we were able to estimate the best possible values of the exponents.
\begin{table}
\caption{The best possible average cluster recovery, in terms of ARI, provided by Ward$_p$ and A-Ward$_{p\beta}$. The ARI's standard deviation and the pre-selected number of clusters, $K^*$, found at the anomalous pattern initialisation step of A-Ward$_{p\beta}$ are also indicated.}
\begin{center}
\tabcolsep=0.11cm
\begin{tabular}{lcccccccccccccccccc}
&\multicolumn{2}{c}{Ward$_p$}&&\multicolumn{5}{c}{A-Ward$_{p \beta}$}\\
\cline{5-9}
&&&&\multicolumn{2}{c}{ARI}&&\multicolumn{2}{c}{$K^*$}\\
\cline{2-3}
\cline{5-6}
\cline{8-9}
&avg&sd&&avg&sd&&avg&sd\\
1000x6-3&0.6568&0.154&&0.7314&0.135&&18.45&3.220\\
1000x6-3 +3NF&0.3193&0.249&&0.6348&0.195&&16.20&3.650\\
1000x6-3 50\%N&0.2831&0.163&&0.4851&0.190&&16.50&3.502\\ \hline
1000x12-6&0.7412&0.148&&0.8066&0.121&&21.25&4.253\\
1000x12-6 +6NF&0.3440&0.212&&0.7467&0.161&&15.90&2.532\\
1000x12-6 50\%N&0.2535&0.191&&0.6138&0.147&&17.10&3.655\\ \hline
1000x20-10&0.9119&0.035&&0.9564&0.021&&22.20&5.988\\
1000x20-10 +10NF&0.4638&0.098&&0.9258&0.025&&27.20&6.118\\
1000x20-10 50\%N&0.2021&0.096&&0.8440&0.042&&23.25&4.833\\
\end{tabular}
\end{center}
\label{Tab:GMM_Baseline}
\end{table}

Table \ref{Tab:GMM_Baseline} shows that the best possible average ARI of Ward$_p$ and A-Ward$_{p\beta}$ decreases when noise is added to the data sets, but not as much as it decreases in the case of traditional Ward (see Table \ref{Tab:GMM_Baseline_Ward_TTWard}). A-Ward$_{p\beta}$ is particularly impressive at the largest structure configuration, 1000x20-10. When 10 noise features are added to data sets (configuration 1000x20-10 +10NF), the average ARI obtained by Ward falls from $0.8998$ to $0.2418$. If instead of adding 10 noise features we substitute 50\% of the cluster-specific data with noise (configuration 1000x20-10 50\%N), the ARI falls even further to $0.1360$. That is a decrease of over six times. Ward$_p$ presents considerable falls of ARI in the same scenario, too. In contrast, the accuracy decrease of Ward$_{p\beta}$ is only about $0.03$ when 10 noise features are added to the data. Furthermore, the average ARI obtained with A-Ward$_{p\beta}$ over the data sets 1000x20-10 + 10NF is nearly twice that of Ward$_p$, and nearly four times that of Ward. The experiments with the data sets 1000x20 50\%N show a very similar trend. The average ARI obtained by A-Ward$_{p\beta}$ is about four times higher than that of Ward$_p$, and about six times higher than that of Ward.

Thus, in an ideal situation of the known best $p$ and $\beta$, A-Ward$_{p\beta}$ is capable of obtaining really good clusterings that are much superior to those yielded by Ward and Ward$_p$. 

\subsection{Estimating the optimal values of the exponents $p$ and $\beta$} 
\label{Sec:Estimating_pb}

To find good values for $p$ and $\beta$ in an unsupervised situation, we opted for the Silhouette width cluster validity index \cite{rousseeuw1987silhouettes} which proved successful in the literature \cite{arbelaitz2012extensive} as well as in our previous experiments \cite{de2016applying,de2015feature,de2015recovering}. The Silhouette width of a partition $S$ is the average Silhouette width of entities $y_i \in Y$, defined as follows:
\begin{equation}
\label{Eq:Silhouette}
Sil(y_i) = \frac{b(y_i)-a(y_i)}{max\{a(y_i), b(y_i)\}},
\end{equation}
where $a(y_i)$ is the average dissimilarity of $y_i \in S_k$ to all other $y_j \in S_k$, and $b(y_i)$ the minimum dissimilarity over all clusters $S_q \in S$, to which $y_i$ is not assigned, of the average dissimilarities to $y_j \in S_q, q\not=k$. Therefore, $-1\leq Sil(y_i) \leq 1$. A $Sil(y_i)$ value near zero indicates that $y_i$ could be assigned to another cluster without much damaging both cluster cohesion and separation. A negative $Sil(y_i)$ suggests that $y_i$'s cluster assignment is damaging to the cluster cohesion and separation, whereas an $Sil(y_i)$ closer to one means the opposite. We can then quantify the validity of the whole clustering $S$ by the Silhouette index, defined as $Sil=1/N\sum_{i \in Y} Sil(y_i)$. 

Table \ref{Tab:GMM_Wardpb} reports the average ARI and standard deviations of Ward$_{p\beta}$, obtained with the estimated values of $p$ and $\beta$, for each of the nine parameter configurations. The exponents $p$ and $\beta$ have been estimated as those corresponding to the highest values of  the average Silhouette width (\ref{Eq:Silhouette}). We have experimented with the Silhouette width validity index measured using the squared Euclidean, Manhattan and Minkowski distances. The exponent of the latter was set to the same value of $p$ that was used in A-Ward$_{p\beta}$.
\begin{table}\small
\caption{Average ARI and its standard deviations for clustering solutions found using A-Ward$_{p\beta}$. The best possible results for this algorithm are presented under the column \textit{Best}. Under \textit{Silhouette}, we present the results for $p$ and $\beta$ estimated using this cluster validity index, with either the squared Euclidean distance, or Manhattan distance, or Minkowski distance. In the latter case, the Minkowski exponent was set to the same value of $p$ that was used in A-Ward$_{p\beta}$.}
\begin{center}
\tabcolsep=0.11cm
\begin{tabular}{lccccccccccc}
&&&&\multicolumn{8}{c}{Silhouette}\\
\cline{5-12}
&\multicolumn{2}{c}{Best}&&\multicolumn{2}{c}{sq. Euclidean}&&\multicolumn{2}{c}{Manhattan}&&\multicolumn{2}{c}{Minkowski}\\
\cline{2-3}
\cline{5-6}
\cline{8-9}
\cline{11-12}
&avg&sd&&avg&sd&&avg&sd&&avg&sd\\
1000x6-3&0.7314&0.135&&0.6476&0.189&&0.6351&0.193&&0.6706&0.170\\
1000x6-3 3NF&0.6348&0.195&&0.1785&0.269&&0.3475&0.299&&0.1838&0.289\\
1000x6-3 50\%N&0.4851&0.190&&0.1285&0.219&&0.1715&0.243&&0.1026&0.199\\
\hline
1000x12-6&0.8066&0.121&&0.7109&0.178&&0.7035&0.183&&0.7200&0.185\\
1000x12-6 6NF&0.7467&0.161&&0.4693&0.237&&0.6279&0.236&&0.5818&0.232\\
1000x12-6 50\%N&0.6138&0.147&&0.2596&0.213&&0.2937&0.237&&0.2592&0.237\\
\hline
1000x20-10&0.9564&0.021&&0.9254&0.035&&0.9216&0.037&&0.9185&0.036\\
1000x20-10 10NF&0.9258&0.025&&0.8585&0.076&&0.8849&0.052&&0.8732&0.044\\
1000x20-10 50\%N&0.8440&0.042&&0.5122&0.211&&0.7271&0.096&&0.6363&0.195\\
\end{tabular}
\end{center}
\label{Tab:GMM_Wardpb}
\end{table}

Table \ref{Tab:GMM_Wardpb} replicates the best possible average ARI values of A-Ward$_{p\beta}$ from Table \ref{Tab:GMM_Baseline}. The results reported in Table \ref{Tab:GMM_Wardpb} show some interesting patterns. Probably the most striking of them is that all the average ARI values obtained by A-Ward$_{p\beta}$ using the estimated values of $p$ and $\beta$ are much better than the average ARI values of the conventional Ward shown in Table \ref{Tab:GMM_Baseline_Ward_TTWard}. The results obtained by A-Ward$_{p\beta}$ are also superior to the best possible results of Ward$_p$ in a number of occasions. This is particularly true for the experiments carried out at greater numbers of clusters: 1000x12-6 6NF, 1000x12-6 50\%N, 1000x20-10 10NF, and 1000x20-10 50\%N. It should be pointed out that, in these experiments, using Manhattan distance for calculation of the Silhouette width index leads to better cluster recovery results overall. It would be fair to say that the results provided by A- Ward$_{p\beta}$, with the exponents $p$ and $\beta$ estimated using the Silhouette cluster validity index, are promising indeed.

\section{Conclusion}

This paper makes two novel contributions to hierarchical clustering. First, we introduced an initialisation method, A-Ward, for hierarchical clustering algorithms. This method generates initial partitions with a sufficiently large number of clusters. Thus, the cluster merging process begins from this partition rather than from a trivial partition composed solely of singletons. The anomalous pattern initialisation method can reduce substantially the time a hierarchical clustering algorithm takes to complete without negatively impacting its cluster recovery ability.

Second, we introduced A-Ward$_{p\beta}$, a novel hierarchical clustering algorithm which can be viewed as an extension of the popular Ward algorithm. Ward$_{p\beta}$ applies a feature weighted version of the Minkowski distance, making it able to detect clusters with shapes other than spherical. The 
feature weights are cluster-specific. They follow the intuitive idea that the relevance of a feature at a particular cluster is inversely proportional to its dispersion within that cluster. Thus, a feature with a low dispersion within a certain cluster has a higher degree of relevance than a feature with a high dispersion. The computation process according to A-Ward$_{p\beta}$ incorporates this concept via the use of cluster specific feature weights. The new algorithm is initialised with our anomalous pattern identification method.

We empirically validated the anomalous pattern initialisation method in the framework of both Ward and Ward$_{p\beta}$ by running a number of simulations with synthetic data sets. We experimented with numerous data sets containing Gaussian clusters, with and without noise added to them. In contrast to our previous experiments, here noise has been added in two different ways: (i) each data set was supplemented with features composed entirely of uniform random values, the number of features added was equal to the half of the number of original features; (ii) cluster specific noise was generated by substituting 50\% of the cluster-specific data fragments by uniform random values.

In our experiments we compared the Ward, A-Ward, Ward$_p$ and A-Ward$_{p\beta}$ algorithms in terms of cluster recovery. To do so, we measured the average Adjusted Rand Index for the obtained clustering solutions found by these algorithms in relation to the known truth. Our main conclusion is that A-Ward$_{p\beta}$ is capable of good cluster recovery in difficult practical situations. It produces superior results to those of Ward and Ward$_p$, especially when data sets are affected by the presence of noise features. This is in fact the case for most real-world data.

Our future research will investigate other methods for estimation of $p$ and $\beta$ as well as further advancements into the problem of evaluation of the true number of clusters using both divisive and agglomerative hierarchical clustering algorithms.

%
%
%
%
\bibliography{references}

\begin{thebibliography}{10}
\expandafter\ifx\csname url\endcsname\relax
  \def\url#1{\texttt{#1}}\fi
\expandafter\ifx\csname urlprefix\endcsname\relax\def\urlprefix{URL }\fi

\bibitem{arbelaitz2012extensive}
O.~Arbelaitz, I.~Gurrutxaga, J.~Muguerza, J.~M. P{\'e}rez, I.~Perona, An
  extensive comparative study of cluster validity indices, Pattern Recognition
  46~(1) (2012) 243–--256.

\bibitem{ball1967clustering}
G.~H. Ball, D.~J. Hall, A clustering technique for summarizing multivariate
  data, Behavioral Science 12~(2) (1967) 153--155.

\bibitem{bezdek1984fcm}
J.~C. Bezdek, R.~Ehrlich, W.~Full, {FCM}: The fuzzy c-means clustering
  algorithm, Computers \& Geosciences 10~(2) (1984) 191--203.

\bibitem{bradley1998refining}
P.~S. Bradley, U.~M. Fayyad, Refining initial points for k-means clustering,
  in: Proceedings of the 15th International Conference on Machine Learning,
  Morgan Kaufmann, San Francisco, USA, 1998, pp. 91--99.

\bibitem{cao2009initialization}
F.~Cao, J.~Liang, G.~Jiang, An initialization method for the k-means algorithm
  using neighborhood model, Computers \& Mathematics with Applications 58~(3)
  (2009) 474--483.

\bibitem{celebi2012deterministic}
M.~E. Celebi, H.~A. Kingravi, Deterministic initialization of the k-means
  algorithm using hierarchical clustering, International Journal of Pattern
  Recognition and Artificial Intelligence 26~(7) (2012) 1250018.

\bibitem{chiang2010intelligent}
M.~M.-T. Chiang, B.~Mirkin, Intelligent choice of the number of clusters in
  k-means clustering: an experimental study with different cluster spreads,
  Journal of Classification 27~(1) (2010) 3--40.

\bibitem{de2015feature}
R.~C. de~Amorim, Feature relevance in {Ward's} hierarchical clustering using
  the {Lp} norm, Journal of Classification 32~(1) (2015) 46--62.

\bibitem{de2015recovering}
R.~C. de~Amorim, C.~Hennig, Recovering the number of clusters in data sets with
  noise features using feature rescaling factors, Information Sciences 324
  (2015) 126--145.

\bibitem{de2016applying}
R.~C. de~Amorim, V.~Makarenkov, Applying subclustering and {Lp} distance in
  weighted k-means with distributed centroids, Neurocomputing 173 (2016)
  700--707.

\bibitem{cordeiro2011minkowski}
R.~C. de~Amorim, B.~Mirkin, Minkowski metric, feature weighting and anomalous
  cluster initializing in k-means clustering, Pattern Recognition 45~(3) (2012)
  1061--1075.

\bibitem{dempster1977maximum}
A.~P. Dempster, N.~M. Laird, D.~B. Rubin, Maximum likelihood from incomplete
  data via the {EM} algorithm, Journal of the Royal Statistical Society. Series
  B (Methodological) (1977) 1--38.

\bibitem{eppstein2000fast}
D.~Eppstein, Fast hierarchical clustering and other applications of dynamic
  closest pairs, Journal of Experimental Algorithmics (JEA) 5 (2000) 1--23.

\bibitem{fraley1998many}
C.~Fraley, A.~E. Raftery, How many clusters? which clustering method? answers
  via model-based cluster analysis, The Computer Journal 41~(8) (1998)
  578--588.

\bibitem{freytag2012efficient}
A.~Freytag, B.~Frohlich, E.~Rodner, J.~Denzler, Efficient semantic segmentation
  with gaussian processes and histogram intersection kernels, in: 21st
  International Conference on Pattern Recognition (ICPR), IEEE, 2012, pp.
  3313--3316.

\bibitem{hubertarabie1985rand}
L.~Hubert, P.~Arabie, Comparing partitions, Journal of Classification 2~(2)
  (1985) 193--218.

\bibitem{jain2010data}
A.~Jain, Data clustering: 50 years beyond k-means, Pattern Recognition Letters
  31~(8) (2010) 651--666.

\bibitem{Juan1982Programme}
J.~Juan, Programme de classification hi{\'e}rarchique par l'algorithme de la
  recherche en cha{\^\i}ne des voisins r{\'e}ciproques, Les Cahiers de
  L'Analyse des Donn{\'e}es 7~(2) (1982) 219--225.

\bibitem{kriegel2009clustering}
H.-P. Kriegel, P.~Kr{\"o}ger, A.~Zimek, Clustering high-dimensional data: A
  survey on subspace clustering, pattern-based clustering, and correlation
  clustering, ACM Transactions on Knowledge Discovery from Data (TKDD) 3~(1)
  (2009) 1--58.

\bibitem{leiva2013warped}
L.~A. Leiva, E.~Vidal, Warped k-means: An algorithm to cluster
  sequentially-distributed data, Information Sciences 237 (2013) 196--210.

\bibitem{Lichman:2013}
M.~Lichman, {UCI} machine learning repository (2013).
\newline\urlprefix\url{http://archive.ics.uci.edu/ml}

\bibitem{macqueen1967some}
J.~MacQueen, Some methods for classification and analysis of multivariate
  observations, in: Proceedings of the Fifth Berkeley Symposium on Mathematical
  Statistics and Probability, vol.~1, California, USA, 1967, pp. 281--297.

\bibitem{makarenkov2001optimal}
V.~Makarenkov, P.~Legendre, Optimal variable weighting for ultrametric and
  additive trees and k-means partitioning: Methods and software, Journal of
  Classification 18~(2) (2001) 245--271.

\bibitem{maldonado2015kernel}
S.~Maldonado, E.~Carrizosa, R.~Weber, Kernel penalized k-means: A feature
  selection method based on kernel k-means, Information Sciences 322 (2015)
  150--160.

\bibitem{milligan1980}
G.~W. Milligan, An examination of the effect of six types of error perturbation
  on fifteen clustering algorithms, Psychometrika 45~(3) (1980) 325--342.
\newline\urlprefix\url{http://dx.doi.org/10.1007/BF02293907}

\bibitem{mirkin2012clustering}
B.~Mirkin, Clustering: A Data Recovery Approach, Computer Science and Data
  Analysis, CRC Press, London, UK, 2012.

\bibitem{monni2009stochastic}
S.~Monni, M.~G. Tadesse, et~al., A stochastic partitioning method to associate
  high-dimensional responses and covariates, Bayesian Analysis 4~(3) (2009)
  413--436.

\bibitem{murtagh1983survey}
F.~Murtagh, A survey of recent advances in hierarchical clustering algorithms,
  The Computer Journal 26~(4) (1983) 354--359.

\bibitem{murtagh1985multidimensional}
F.~Murtagh, Multidimensional clustering algorithms, Compstat Lectures, Vienna:
  Physika Verlag, 1985.

\bibitem{murtagh2014ward}
F.~Murtagh, P.~Legendre, {Ward's} hierarchical agglomerative clustering method:
  which algorithms implement {Ward's} criterion?, Journal of Classification
  31~(3) (2014) 274--295.

\bibitem{pena1999empirical}
J.~M. Pena, J.~A. Lozano, P.~Larranaga, An empirical comparison of four
  initialization methods for the k-means algorithm, Pattern Recognition Letters
  20~(10) (1999) 1027--1040.

\bibitem{r_stats}
{R Core Team}, The {R} stats package version 3.4.0 (2013).
\newline\urlprefix\url{https://stat.ethz.ch/R-manual/R-devel/library/stats/html/00Index.html}

\bibitem{rousseeuw1987silhouettes}
P.~J. Rousseeuw, Silhouettes: a graphical aid to the interpretation and
  validation of cluster analysis, Journal of Computational and Applied
  Mathematics 20 (1987) 53--65.

\bibitem{steinbach2000comparison}
M.~Steinbach, G.~Karypis, V.~Kumar, et~al., A comparison of document clustering
  techniques, in: KDD Workshop on Text Mining, Boston, 2000, pp. 525--526.

\bibitem{steinley2006k}
D.~Steinley, K-means clustering: A half-century synthesis, British Journal of
  Mathematical and Statistical Psychology 59~(1) (2006) 1--34.

\bibitem{su2007search}
T.~Su, J.~G. Dy, In search of deterministic methods for initializing k-means
  and gaussian mixture clustering, Intelligent Data Analysis 11~(4) (2007)
  319--338.

\bibitem{tan2006introduction}
P.-N. Tan, M.~Steinbach, V.~Kumar, et~al., Introduction to data mining, vol.~1,
  Pearson Addison Wesley Boston, 2006.

\bibitem{matlab_stats}
{The MathWorks, Inc.}, Matlab and statistics toolbox release 2012b (2012).
\newline\urlprefix\url{http://uk.mathworks.com/products/statistics/}

\bibitem{ward1963hierarchical}
J.~H. Ward~Jr, Hierarchical grouping to optimize an objective function, Journal
  of the American Statistical Association 58~(301) (1963) 236--244.

\bibitem{wilcox2014hierarchical}
D.~Wilcox, T.~Gebbie, Hierarchical causality in financial economics, Social
  Science Research Network, 2544327.
\newline\urlprefix\url{https://dx.doi.org/10.2139/ssrn.2544327}

\bibitem{Clustan}
D.~Wishart, Clustan (1998).
\newline\urlprefix\url{http://www.clustan.com/}

\bibitem{zadeh1965fuzzy}
L.~A. Zadeh, Fuzzy sets, Information and Control 8~(3) (1965) 338--353.

\end{thebibliography}
\end{document}